\documentclass[letterpaper, 10 pt, conference]{ieeeconf}%

\IEEEoverridecommandlockouts%
                                                          
\usepackage{cite}
\usepackage{amsmath,amssymb,amsfonts}
\usepackage{graphicx}
\usepackage{textcomp}
\usepackage{xcolor}
\usepackage{url}
\usepackage{algorithm,algpseudocode}
\usepackage{booktabs}
\usepackage{siunitx}
\usepackage{tikz}
\usepackage{pgfplots}
\pgfplotsset{compat=1.18}
\usepackage{balance}
\usetikzlibrary{arrows.meta,positioning,shapes.geometric,calc}

\def\BibTeX{{\rm B\kern-.05em{\sc i\kern-.025em b}\kern-.08em
    T\kern-.1667em\lower.7ex\hbox{E}\kern-.125emX}}

\begin{document}

\title{Decentralized Scalable Exploration via Emergent Adaptive L\'evy Walks on Minimal-Sensing Platforms\\%
}

\author{Wai Lun Leong$^{1}$ and Teo Swee Huat Rodney$^{2}$%
\thanks{Additional material including the open-source code is available at \protect\url{https://github.com/williamleong/sdlw}.}%
\thanks{$^{1,2}$Wai Lun Leong and Teo Swee Huat Rodney are with the National University of Singapore, Singapore {\tt\small \{william.leong, tsltshr\}@nus.edu.sg}}%
}

\maketitle

\begin{abstract}
Efficient autonomous exploration with palm-sized nano-UAVs remains challenging due to severe limitations in sensing, computation, and flight endurance. We present a lightweight sensor-driven L\'evy walk (SDLW) controller for aerial robots weighing under \SI{50}{\gram} equipped with sparse local sensing. The method combines discrete L\'evy step-length sampling with a sensor-reactive heading policy using directional range measurements. Each robot independently samples its L\'evy exponent from a uniform prior to diversify exploration without inter-robot communication for exploration control. Each robot then selects headings using a von Mises distribution that biases motion toward open directions while preserving superdiffusive exploration properties. The controller operates at constant computational cost, enabling scalable multi-UAV exploration. Simulation results show coverage improvements of 79.6\% in open arenas, 43.1\% in rooms-and-corridors layouts, and 13.6\% in cluttered environments, with collision reductions of 13.0\%, 7.1\%, and 1.4\% respectively, relative to a uniform-heading L\'evy walk baseline. This work provides a practical framework for scalable multi-robot exploration on minimal-sensing, resource-constrained nano-UAVs.
\end{abstract}

\section{Introduction}
Palm-sized nano-UAVs offer a critical capability for time-sensitive search-and-rescue missions in hazardous post-disaster environments, such as structurally unstable buildings, confined spaces with toxic atmospheres, or rubble-filled passages where human access is dangerous or impossible, yet autonomous exploration remains challenging due to severe constraints in onboard sensing, computation, and flight endurance. These constraints often rule out SLAM-centric or optimization-heavy planning. Recent work has demonstrated that lightweight SLAM can run onboard nano-UAVs using minimal time-of-flight (ToF) sensors and Cortex-M microcontrollers \cite{friess2024fully,niculescu2025ultra}, but such systems still require careful algorithm optimization; for missions where only coverage rather than detailed mapping is required, simpler reactive approaches remain attractive. In such regimes, reactive controllers that rely only on local measurements are appealing: they incur constant computational cost, avoid map maintenance, and maintain constant-time complexity with multiple agents without inter-robot communication. Among minimalist search processes, L\'evy walks provide a principled balance between short local probes and occasional long relocations via heavy-tailed step lengths that yield superdiffusive dispersion; the model and its broad relevance have been surveyed in depth in physics and biology \cite{zaburdaev2015levy}.

Recent theory on the discrete two-dimensional grid $\mathbb{Z}^2$ shows that for parallel search there is an optimal exponent $\alpha \in (2,3)$ that depends on team size $k$ and search scale $\ell$. Sampling each agent's exponent independently and uniformly in $(2,3)$ achieves near-optimal parallel search performance over wide ranges of $k$ and $\ell$ (within polylog factors)
. The same line of work formalizes the discrete L\'evy walk with integer jump distances distributed as $P(d{=}0)=1/2$ and $P(d{=}i) \propto i^{-\alpha}$ for $i \ge 1$ \cite{clementi2021search}. Preserving this discrete formulation helps keep execution lightweight while remaining aligned with theoretical guarantees.

Translating these ideas to resource-constrained robots has taken several directions. One approach biases headings using \emph{a priori} target densities, selecting directions from a von Mises distribution centered on the gradient $\nabla P$ of a known field; this bias improves search when a reliable prior exists, but it presupposes access to that prior, which nano-UAVs commonly lack \cite{marthaler2004levy}. A second approach designs information-correlated L\'evy walks in which each step’s heading maximizes predicted mutual information with an occupancy map under construction; this improves mapping but requires maintaining a map and evaluating information gain at each step \cite{ramachandran2020information}. A third approach modifies L\'evy motion using potential fields for obstacle avoidance and inter-robot spacing \cite{sutantyo2010multi}. Related work also explores reactive L\'evy-like dispersion without communication to speed up coverage, and collective L\'evy-walk mechanisms to preserve heavy-tailed properties at larger swarm sizes \cite{beal2013superdiffusive, khaluf2018collective}.

\paragraph*{Gap and Goal}
We propose a fully reactive, map-free controller combining discrete Lévy steps with sensor-driven von Mises heading selection. A sensor-weighted resultant from four directional ranges biases motion toward open space via a resultant-strength-to-concentration mapping, preserving superdiffusion. The design requires $O(1)$ computation, no inter-robot communication, and maintains constant-time complexity across team sizes.

\subsection{Problem Statement and Assumptions}

We consider a team of $k$ palm-sized nano-UAVs tasked with coverage-oriented exploration of a bounded, cluttered 2-D domain. This work targets platforms such as the Bitcraze Crazyflie nano-UAV (mass around \SIrange{27}{33}{\gram})\cite{giernacki2017crazyflie,mcguire2019minimal}, which exemplifies the sub-\SI{50}{\gram} minimal-sensing constraint. In prior indoor exploration deployments, Crazyflie setups use a \emph{Multi-ranger deck} for four directional distance readings and a \emph{Flow deck} for optical-flow motion estimation \cite{mcguire2019minimal}.

\emph{Sensing.} Each UAV acquires four directional distance measurements in its local frame (front, left, right, back) per control cycle. This sparse sensing provides essential environmental information while maintaining a minimal computational burden. Critically, there is no global localization, no map building, and no access to priors such as target distributions or occupancy models. Each agent must decide reactively, based exclusively on current local distance measurements.

\emph{Computation.} Onboard compute is severely constrained. All control updates must execute in constant time $O(1)$ with kilobyte-scale memory. This rules out map maintenance, global optimization, and complex state estimation, placing strong emphasis on lightweight, reactive controllers.

\emph{Communication and coordination.} Search behavior is fully decentralized and does not rely on inter-robot communication for exploration control. Lightweight message passing may still be available for mission-critical events (e.g., safety or failsafe alerts), but it is not used for search decisions. This preserves scalable emergent multi-robot behavior while avoiding communication-dependent exploration logic.

\subsection{Contributions}
This work contributes:
\begin{enumerate}
  \item \textbf{Sensor-Driven Lévy-Walk search controller for nano-UAVs.} We propose a map-free, constant-time controller using only sparse local directional ranging to stochastically bias headings via von Mises distribution while executing discrete Lévy steps.
  \item \textbf{Theory-informed exponent randomization.} Following established results that $\alpha \sim \mathrm{Uniform}(2,3)$ yields near-optimal parallel search performance, we realize this strategy on minimal-sensing nano-UAVs.
  \item \textbf{Sensor-adaptive stochastic steering.} We introduce a resultant-strength-to-concentration mapping that adapts von Mises concentration to local structure, preserving superdiffusive behavior while reducing collisions.
  \item \textbf{Comprehensive simulation evaluation.} Evaluation across three arena types and five team sizes demonstrates 13\%--65\% coverage gains and 1\%--29\% collision reduction relative to baseline; simulation code is released to enable reproducibility and further exploration.
  \item \textbf{Design for nano-UAV resource constraints.} We provide a constant-time implementation relying only on four directional readings and lightweight sampling suitable for sub-\SI{50}{\gram} platforms.
\end{enumerate}

\subsection{Related Works}
\textbf{Theory of Lévy search.} Clementi et al.\ establish that per-agent $\alpha \sim \mathrm{Uniform}(2,3)$ is near-optimal for parallel search on $\mathbb{Z}^2$ across unknown team sizes and scales. We adopt this result and validate its feasibility on resource-constrained platforms \cite{clementi2021search}.

\textbf{Biasing directions: prior fields vs.\ reactive sensing.} Marthaler et al.\ bias headings using a known target-density gradient and a von Mises directional model, but their method requires priors \cite{marthaler2004levy}. Ramachandran et al.\ maximize mutual information for mapping, which requires maintaining a map and evaluating information gain each step \cite{ramachandran2020information}. Our reactive sensor-weighted von~Mises achieves directional bias without priors or maps, suitable for constant-time nano-UAV control.

\textbf{Potential-field Lévy hybrids and reactive dispersion.} Works combining Lévy motion with potential fields offer practical collision avoidance but rely on deterministic steering. Our stochastic von~Mises policy maintains controlled randomness tunable via $\kappa$, better preserving superdiffusive properties. Reactive Lévy-like dispersion results support scalability without communication \cite{sutantyo2010multi,beal2013superdiffusive,khaluf2018collective}.

\textbf{Minimal navigation for nano-UAV swarms.} McGuire et al.\ demonstrate exploration with Crazyflie swarms using wall-following and RSSI-homing (SGBA), employing deterministic bug-algorithm steering \cite{mcguire2019minimal}. Our work differs from this by using reactive stochastic heading selection throughout exploration, preserving Lévy walk superdiffusive properties without return-to-home infrastructure, focusing on coverage-oriented search rather than outbound-return missions.

The above approaches lack a fully reactive, map-free, minimal-sensing controller that simultaneously (i) preserves superdiffusive Lévy properties, (ii) exploits per-agent exponent randomization, and (iii) adapts heading reactively to local geometry in constant time. We address this gap by combining discrete Lévy steps with a sensor-driven von~Mises heading policy that biases motion toward open space while retaining controlled randomness. The controller requires no inter-robot communication and no shared map, enabling implicit cooperation through decentralized motion rather than explicit coordination.

\section{Method}
This section formalizes the Sensor-Driven Lévy-Walk (SDLW) controller as a constant-time reactive pipeline. Each control cycle uses four local range measurements to estimate open directions, samples a discrete Lévy segment length to preserve heavy-tailed exploration, and selects a stochastic heading from a sensor-conditioned von Mises distribution. The resulting command is executed by a two-state \textsf{Rotate}/\textsf{Move} controller.

The subsections follow this pipeline. Section~II-A defines the minimal agent and sensing model. Section~II-B specifies the discrete Lévy step-length sampler used to generate exploratory motion segments. Section~II-C defines the sensor-weighted heading policy that biases those segments toward open space while retaining stochasticity. Section~II-D then combines these components into the full reactive controller.

\subsection{Agent and Sensing Model}
\begin{figure}[tbp]
\centering
\begin{tikzpicture}[
  scale=1.0,
  every node/.style={font=\small},
  >=Latex
]
\draw[fill=gray!20, very thick] (0,0) circle (0.3);
\node at (0,0) {\textsf{R}};

\draw[thick, blue!70, ->] (0,0) -- (0,1.2) node[above, font=\small\bfseries] {$r_F$};
\draw[thick, blue!70, ->] (0,0) -- (1.2,0) node[right, font=\small\bfseries] {$r_R$};
\draw[thick, blue!70, ->] (0,0) -- (-1.2,0) node[left, font=\small\bfseries] {$r_L$};
\draw[thick, blue!70, ->] (0,0) -- (0,-1.2) node[below, font=\small\bfseries] {$r_B$};

\draw[dashed, gray, thin] (0,0) circle (1.3);
\end{tikzpicture}
\caption{Sensor geometry (top-down view). The robot acquires four orthogonal distance measurements ($r_F, r_L, r_R, r_B$) in its local frame, corresponding to front, left, right, and back directions.}
\label{fig:sensor-geometry}
\end{figure}
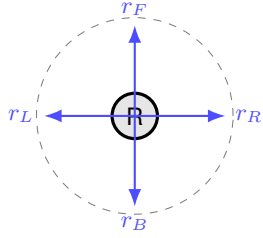

Consider $k$ identical palm-sized nano-UAVs operating in a bounded, cluttered 2-D domain. Each agent $i\!\in\!\{1,\dots,k\}$ maintains a planar pose $(x_i,y_i,\psi_i)$, where $\psi_i$ is the yaw (heading). Per control cycle, the agent acquires a fixed set of directional distance measurements
\[
\mathcal{R}_i \triangleq \{r_F, r_L, r_R, r_B\}\!,
\]
corresponding to front $(0)$, left $(+\pi/2)$, right $(-\pi/2)$, and back $(\pi)$ directions in the agent frame (Fig.~\ref{fig:sensor-geometry}). There is no global localization and no mapping; search decisions rely only on $\mathcal{R}_i$ and constant-size state, without explicit coordination.

\subsection{Discrete L\'evy Step-Length Sampling}
Each agent performs motion in \emph{segments} (``jumps'') of length $d\!\ge\!0$. Let the agent’s L\'evy exponent $\alpha_i$ be sampled once at initialization,
\[
\alpha_i \sim \mathrm{Uniform}(2,3),
\]
to diversify search across the team without explicit coordination. Given bounds $d_{\min}\!>\!0$ and $d_{\max}\!>\!d_{\min}$, we adopt the discrete law:
\begin{align}
\Pr[d=0] &= \tfrac{1}{2}, \label{eq:stay}\\
\Pr[d=j] &= \frac{j^{-\alpha_i}}{\sum_{\ell=j_{\min}}^{j_{\max}} \ell^{-\alpha_i}}, \quad j \in \{j_{\min},\dots,j_{\max}\}, \label{eq:levy}
\end{align}
where $j_{\min}\!=\!\lceil d_{\min}\rceil$ and $j_{\max}\!=\!\lfloor d_{\max}\rfloor$.

We retain $\Pr[d=0]=1/2$ exactly to match the discrete formulation in \cite{clementi2021search} rather than introduce an extra design parameter. Operationally, a $d=0$ draw yields a rotate-only cycle: the agent reorients to a new heading without forward motion, then samples a fresh step length $d$ for the subsequent move phase. The same obstacle-threshold checks apply during rotation, so small axial or lateral nudges can occur if any range is below the safety threshold. Algorithm~\ref{alg:discrete-levy} provides the complete sampling procedure, which executes in constant time due to the fixed support size.

\begin{algorithm}[t]
\caption{Discrete L\'evy Step-Length Sampler}
\label{alg:discrete-levy}
\begin{algorithmic}[1]
\Require $\alpha \in (2,3)$,\ $d_{\min}>0$,\ $d_{\max}>d_{\min}$
\State $j_{\min} \leftarrow \lceil d_{\min}\rceil$;\quad $j_{\max} \leftarrow \lfloor d_{\max}\rfloor$
\State With probability $1/2$, \Return $0$
\State $Z \leftarrow \sum_{\ell=j_{\min}}^{j_{\max}} \ell^{-\alpha}$
\State Sample $j \in \{j_{\min},\dots,j_{\max}\}$ with $\Pr[j]=j^{-\alpha}/Z$
\State \Return $j$
\end{algorithmic}
\end{algorithm}

\subsection{Sensor-Driven Stochastic Heading Selection}

\begin{figure}[tbp]
\centering
\begin{tabular}{@{}c@{\hspace{10mm}}c@{}}
\textbf{(a) Corridor} & \textbf{(b) Junction} \\
\raisebox{-3.5cm}{\begin{tikzpicture}[scale=1.2, every node/.style={font=\small}, >=Latex]
\draw[very thick, black] (-1.0, -1.2) -- (-1.0, 2.0);%
\draw[very thick, black] (0.8, -1.2) -- (0.8, 2.0);%
\draw[fill=black!60, draw=black, very thick] (-0.3, -1.2) rectangle (0.3, -0.8);%
\node[font=\scriptsize, align=center] at (0, -1.35) {obstacle};

\draw[fill=gray!30] (0,0) circle (0.15);

\draw[thick, blue, ->] (0,0) -- (0, 2.0) node[above, font=\small] {$w_F$};
\draw[thick, green, ->] (0,0) -- (-1.0, 0) node[left, font=\small] {$w_L$};
\draw[thick, green, ->] (0,0) -- (0.8, 0) node[right, font=\small] {$w_R$};
\draw[thick, orange, ->] (0,0) -- (0, -0.8) node[above left, font=\small] {$w_B$};

\draw[very thick, red, ->] (0,0) -- (-0.3, 1.9) node[above left, font=\small] {$\mathbf{v}$};
\node[below, font=\footnotesize, gray] at (0, -1.6) {Higher $o$};
\end{tikzpicture}} &
\raisebox{-3.5cm}{\begin{tikzpicture}[scale=1.2, every node/.style={font=\small}, >=Latex]
\draw[very thick, black] (-1.0, -1.2) -- (-1.0, 0.1);%
\draw[very thick, black] (-1.0, 0.1) -- (-1.5, 0.1);%
\draw[very thick, black] (0.75, -1.2) -- (0.75, 1.4);%
\draw[very thick, black] (-1.5, 1.4) -- (0.75, 1.4);%
\draw[fill=black!60, draw=black, very thick] (-0.25, -0.7) rectangle (0.25, -0.2);%
\node[font=\scriptsize, align=center] at (0, -0.85) {obstacle};

\draw[fill=gray!30] (0, 0.85) circle (0.15);

\draw[thick, blue, ->] (0, 0.85) -- (0, 1.4) node[above, font=\small] {$w_F$};
\draw[thick, green, ->] (0, 0.85) -- (-1.5, 0.85) node[left, font=\small] {$w_L$};
\draw[thick, green, ->] (0, 0.85) -- (0.75, 0.85) node[right, font=\small] {$w_R$};
\draw[thick, orange, ->] (0, 0.85) -- (0, -0.2) node[above left, font=\small] {$w_B$};

\draw[very thick, red, ->] (0, 0.85) -- (-0.45, 0.65) node[below left, font=\small] {$\mathbf{v}$};
\node[below, font=\footnotesize, gray] at (0, -1.6) {Lower $o$};
\end{tikzpicture}}
\end{tabular}
\caption{Sensor-weighted resultants in a corridor and a junction. The corridor example has a higher normalized resultant strength $o$ than the junction example, indicating greater directional imbalance. Vectors are not to scale.}
\label{fig:sensor-resultant}
\end{figure}
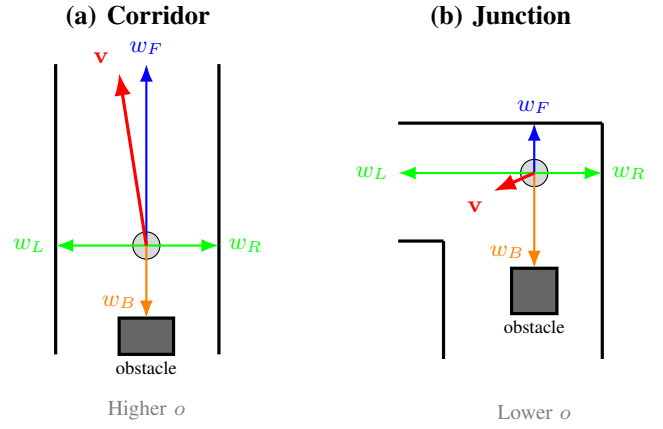

Let the four sensor directions in the world frame be
\[
\Theta \triangleq \{\psi_i,\, \psi_i+\tfrac{\pi}{2},\, \psi_i-\tfrac{\pi}{2},\, \psi_i+\pi\}.
\]
This sensing architecture matches platforms such as the Bitcraze Crazyflie, whose \emph{Multi-ranger deck} provides exactly these four orthogonal distance measurements. We compute a \emph{sensor-weighted resultant} that favors open directions and mildly penalizes moving backward. Define per-direction multipliers $\beta\!=\!\{1,1,1,\beta_B\}$ with $0\!\le\!\beta_B\!\le\!1$ (backward penalty), and weights
\[
w_s \triangleq \beta_s\, r_s^2,\quad s\in\{F,L,R,B\}.
\]
The resultant vector is
\[
\mathbf{v} \triangleq \sum_{s} w_s \begin{bmatrix}\cos \Theta_s\\ \sin \Theta_s\end{bmatrix}, \qquad
W \triangleq \sum_s w_s,
\]
with mean direction $\mu \triangleq \mathrm{atan2}(v_y,v_x)$ and \emph{normalized resultant strength} (see Fig.~\ref{fig:sensor-resultant} for examples)

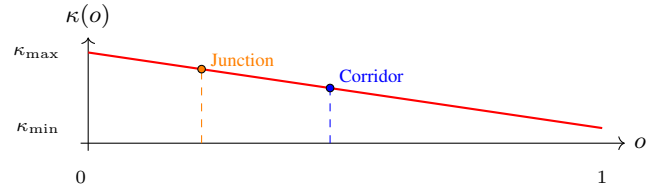
\begin{figure}[tbp]
\centering
\begin{tikzpicture}[every node/.style={font=\small}]
\draw[->] (-0.1, 0) -- (7.1, 0) node[right] {$o$};
\draw[->] (0, -0.1) -- (0, 1.4) node[above] {$\kappa(o)$};

\draw[thick, red] (0, 1.2) -- (6.8, 0.2);

\draw[dashed, orange, thin] (1.50, 0) -- (1.50, 0.98);
\draw[fill=orange, circle] (1.50, 0.98) circle (0.05);
\node[above right, font=\scriptsize, orange] at (1.50, 0.9) {Junction};

\draw[dashed, blue, thin] (3.20, 0) -- (3.20, 0.73);
\draw[fill=blue, circle] (3.20, 0.73) circle (0.05);
\node[below right, font=\scriptsize, blue] at (3.20, 1.10) {Corridor};

\node[below, font=\scriptsize] at (-0.1, -0.25) {$0$};
\node[below, font=\scriptsize] at (6.8, -0.25) {$1$};
\node[left, font=\scriptsize] at (-0.25, 0.2) {$\kappa_{\min}$};
\node[left, font=\scriptsize] at (-0.25, 1.2) {$\kappa_{\max}$};
\end{tikzpicture}
\caption{Resultant-strength-to-concentration mapping. The linear mapping in \eqref{eq:kappa} decreases von Mises concentration as $o$ increases: higher $o$ produces broader angular sampling, while lower $o$ produces narrower sampling around $\mu$.}
\label{fig:kappa-mapping}
\end{figure}

\[
o \triangleq \frac{\|\mathbf{v}\|}{W+\varepsilon}, \quad o\in[0,1],
\]
where $\varepsilon\!>\!0$ is a small numerical guard that keeps the denominator nonzero in degenerate cases, such as when all usable range measurements are zero. The metric $o$ measures normalized directional imbalance in the weighted range measurements. High values indicate a dominant resultant direction, while low values indicate substantial cancellation between opposing contributions.

We map normalized resultant strength to the von~Mises concentration using a decreasing linear function:
\begin{equation}
\kappa(o) \triangleq \kappa_{\max} - (\kappa_{\max}-\kappa_{\min})\, o,
\label{eq:kappa}
\end{equation}
with constants $0\!<\!\kappa_{\min}\!\le\!\kappa_{\max}$ (illustrated in Fig.~\ref{fig:kappa-mapping}). A stronger normalized resultant produces broader angular sampling around the mean direction $\mu$, whereas a weaker resultant produces more concentrated sampling. This empirical heading-selection heuristic adapts angular spread to the relative range measurements and robot orientation.

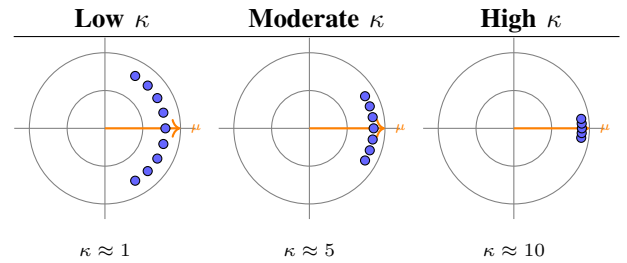
\begin{figure}[tbp]
\centering
\begin{tabular}{@{}c@{\hspace{1mm}}c@{\hspace{1mm}}c@{}}
\textbf{Low }$\kappa$ & \textbf{Moderate }$\kappa$ & \textbf{High }$\kappa$ \\
\hline
\raisebox{-2.5cm}{\begin{tikzpicture}[scale=1.0, every node/.style={font=\scriptsize}]
\draw[gray, thin] (0,0) circle (0.5);
\draw[gray, thin] (0,0) circle (1.0);
\draw[gray, thin] (0,0) -- (1.2, 0);
\draw[gray, thin] (0,0) -- (0, 1.2);
\draw[gray, thin] (0,0) -- (-1.2, 0);
\draw[gray, thin] (0,0) -- (0, -1.2);
\draw[thick, orange, ->] (0,0) -- (1.0, 0) node[right, font=\tiny] {$\mu$};
\foreach \angle in {-60, -45, -30, -15, 0, 15, 30, 45, 60}{
  \draw[fill=blue!60, circle] (\angle:0.8) circle (0.06);
}
\node[below] at (0, -1.4) {$\kappa \approx 1$};
\end{tikzpicture}} &
\raisebox{-2.5cm}{\begin{tikzpicture}[scale=1.0, every node/.style={font=\scriptsize}]
\draw[gray, thin] (0,0) circle (0.5);
\draw[gray, thin] (0,0) circle (1.0);
\draw[gray, thin] (0,0) -- (1.2, 0);
\draw[gray, thin] (0,0) -- (0, 1.2);
\draw[gray, thin] (0,0) -- (-1.2, 0);
\draw[gray, thin] (0,0) -- (0, -1.2);
\draw[thick, orange, ->] (0,0) -- (1.0, 0) node[right, font=\tiny] {$\mu$};
\foreach \angle in {-30, -20, -10, 0, 10, 20, 30}{
  \draw[fill=blue!60, circle] (\angle:0.85) circle (0.06);
}
\node[below] at (0, -1.4) {$\kappa \approx 5$};
\end{tikzpicture}} &
\raisebox{-2.5cm}{\begin{tikzpicture}[scale=1.0, every node/.style={font=\scriptsize}]
\draw[gray, thin] (0,0) circle (0.5);
\draw[gray, thin] (0,0) circle (1.0);
\draw[gray, thin] (0,0) -- (1.2, 0);
\draw[gray, thin] (0,0) -- (0, 1.2);
\draw[gray, thin] (0,0) -- (-1.2, 0);
\draw[gray, thin] (0,0) -- (0, -1.2);
\draw[thick, orange, ->] (0,0) -- (1.0, 0) node[right, font=\tiny] {$\mu$};
\foreach \angle in {-8, -4, 0, 4, 8}{
  \draw[fill=blue!60, circle] (\angle:0.90) circle (0.06);
}
\node[below] at (0, -1.4) {$\kappa \approx 10$};
\end{tikzpicture}} \\
\end{tabular}
\caption{Schematic illustration of angular spread at increasing von Mises concentration. Blue dots illustrate headings around the mean direction $\mu$, directed to the right (orange arrow). Low $\kappa$ permits broader angular sampling; high $\kappa$ concentrates sampling around $\mu$.}
\label{fig:von-mises-concentration}
\end{figure}

The heading command $\theta$ for the next segment is sampled as
\[
\theta \sim \mathrm{VonMises}\big(\mu,\, \kappa(o)\big),
\]
where the effect of varying $\kappa$ on the heading distribution is shown in Fig.~\ref{fig:von-mises-concentration}. The complete sensor-driven heading selection procedure is given in Algorithm~\ref{alg:vm-heading}.

\begin{algorithm}[t]
\caption{Sensor-Driven von Mises Heading}
\label{alg:vm-heading}
\begin{algorithmic}[1]
\Require Ranges $(r_F,r_L,r_R,r_B)$, heading $\psi$, params $(\beta_B,\kappa_{\min},\kappa_{\max},\varepsilon)$
\State $\Theta \leftarrow \{\psi,\ \psi+\tfrac{\pi}{2},\ \psi-\tfrac{\pi}{2},\ \psi+\pi\}$
\State $\beta \leftarrow \{1,1,1,\beta_B\}$;\quad $R \leftarrow \{r_F,r_L,r_R,r_B\}$
\State $w_s \leftarrow \beta_s\,R_s^2$;\quad $W \leftarrow \sum_s w_s$;\quad $\mathbf{v} \leftarrow \sum_s w_s[\cos\Theta_s,\sin\Theta_s]^\top$
\State $\mu \leftarrow \mathrm{atan2}(v_y,v_x)$;\quad $o \leftarrow \|\mathbf{v}\|/(W+\varepsilon)$
\State $\kappa \leftarrow \kappa_{\max}-(\kappa_{\max}-\kappa_{\min})\,\mathrm{clip}(o,0,1)$
\State \Return $\theta \sim \mathrm{VonMises}(\mu,\kappa)$
\end{algorithmic}
\end{algorithm}

\subsection{Control State Machine}

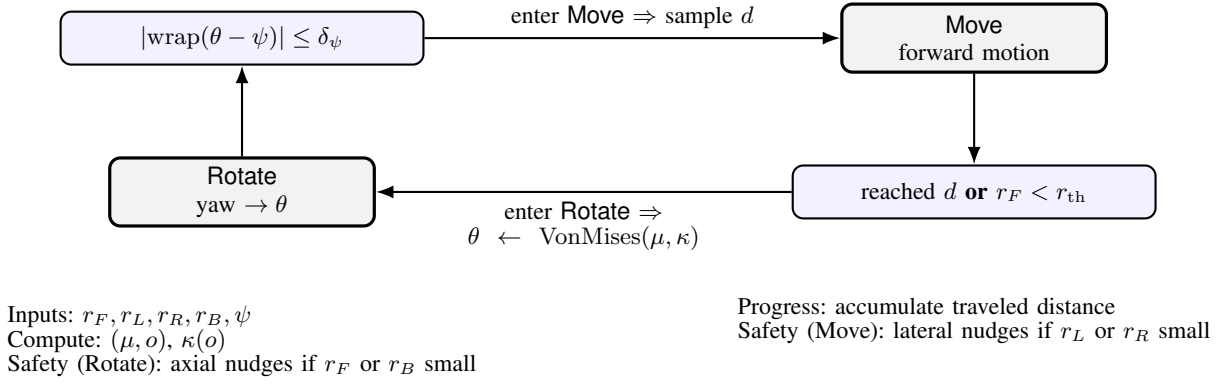
\begin{figure*}[t]
\centering
\begin{tikzpicture}[
  every node/.style={font=\small},
  state/.style={rounded corners, draw, very thick, fill=gray!10, align=center, minimum width=35mm, minimum height=9mm},
  cond/.style={draw, rounded corners, fill=blue!5, align=center, minimum width=48mm, minimum height=7mm},
  >=Latex, thick
]
\node[cond] (c1) {$|\mathrm{wrap}(\theta-\psi)| \le \delta_\psi$};
\node[state, below=12mm of c1] (rotate) {\textsf{Rotate}\\[1pt]yaw $\rightarrow \theta$};
\node[state, right=55mm of c1] (move) {\textsf{Move}\\[1pt]forward motion};
\node[cond, below=12mm of move] (c2) {reached $d$ \textbf{or} $r_F < r_{\mathrm{th}}$};

\draw[->] (rotate) -- (c1);
\draw[->] (c1) -- node[above, align=center, text width=36mm, pos=0.5]{enter \textsf{Move} $\Rightarrow$ sample $d$} (move);
\draw[->] (move) -- (c2);
\draw[->] (c2) -- node[below, align=center, text width=50mm, pos=0.5]{enter \textsf{Rotate} $\Rightarrow$ $\theta \leftarrow \mathrm{VonMises}(\mu,\kappa)$} (rotate);

\node[below=9mm of rotate, align=left] (noteL) {Inputs: $r_F,r_L,r_R,r_B, \psi$\\Compute: $(\mu,o)$,\ $\kappa(o)$\\Safety (Rotate): axial nudges if $r_F$ or $r_B$ small};
\node[below=9mm of c2, align=left] (noteR) {Progress: accumulate traveled distance\\Safety (Move): lateral nudges if $r_L$ or $r_R$ small};

\end{tikzpicture}
\caption{Two-state reactive controller. Agents alternate between \textsf{Rotate}, where heading error is reduced via yaw, and \textsf{Move}, where forward motion progresses until distance $d$ is reached or an obstacle is detected ($r_F<r_{\mathrm{th}}$). The sensor-driven von~Mises policy recomputes heading on each \textsf{Rotate} entry, adapting concentration $\kappa(o)$ to local geometry.}
\label{fig:state-machine}
\end{figure*}

Fig.~\ref{fig:state-machine} depicts the two-state reactive controller structure.

We employ a two-state controller per agent: \textsf{Rotate} $\rightarrow$ \textsf{Move}. Let $\delta_\psi$ be the heading tolerance, and $\tau$ the control period. Controllers initialize in \textsf{Rotate} with $\theta=0$ and sample the first step length $d$ at the first \textsf{Rotate}$\rightarrow$\textsf{Move} transition.

\paragraph*{Rotate}
If $|\mathrm{wrap}(\theta-\psi_i)|>\delta_\psi$, apply a saturated yaw rate toward $\theta$. During rotation, all four ranges are checked and small axial or lateral nudges are applied when any range falls below the safety threshold to create immediate clearance.

\paragraph*{Move}
Advance at constant speed until either (i) the traveled distance reaches $d$, or (ii) an imminent obstacle is detected ($r_F$ below threshold). In either case, switch to \textsf{Rotate} and resample heading via the von~Mises policy. On the next \textsf{Rotate}$\rightarrow$\textsf{Move} transition, sample a new $d$ via \eqref{eq:stay}-\eqref{eq:levy}.

Algorithm~\ref{alg:controller} presents the complete reactive L\'evy-walk controller integrating discrete step sampling, sensor-driven heading selection, and the two-state control structure.

\begin{algorithm}
\caption{Reactive L\'evy-Walk Controller (per agent, per cycle)}
\label{alg:controller}
\begin{algorithmic}[1]
\State \textbf{if} initialization \textbf{then} sample $\alpha_i \sim \mathrm{Uniform}(2,3)$
\If{state = \textsf{Rotate}}
   \If{$|\mathrm{wrap}(\theta-\psi)| \le \delta_\psi$}
      \State $d \leftarrow$ sampleDiscreteLevy$(\alpha_i, d_{\min}, d_{\max})$ \Comment{Alg.~\ref{alg:discrete-levy}}
      \State reset distance accumulator; state $\leftarrow$ \textsf{Move}
   \Else
    \State apply yaw toward $\theta$; axial or lateral nudges if any range is small
   \EndIf
\ElsIf{state = \textsf{Move}}
   \State update traveled distance from odometry/estimate
   \If{(traveled $\ge d$) \textbf{ or } ($r_F < r_{\mathrm{th}}$)}
       \State $(\mu,o)\leftarrow$ sensorResultant$(r_F,r_L,r_R,r_B,\psi)$
       \State $\kappa \leftarrow \kappa(o)$; $\theta \leftarrow \mathrm{sampleVonMises}(\mu,\kappa)$ \Comment{Alg.~\ref{alg:vm-heading}}
       \State state $\leftarrow$ \textsf{Rotate}
     \Else
       \State apply forward velocity; lateral nudges if $r_L$ or $r_R$ small
     \EndIf
\EndIf
\end{algorithmic}
\end{algorithm}

\subsection{Collision Handling and Safety}
We adopt conservative thresholds on front/side ranges for emergency slowdown/stop and small nudges that push away from close obstacles. The concentration mapping controls angular spread, while the range-threshold checks and corrective nudges provide reactive obstacle handling. Backward motion is discouraged by $\beta_B<1$ but permitted when necessary.

\subsection{Implementation for Nano-UAVs}
The control design is amenable to efficient execution on resource-constrained microcontrollers. Computation employs fixed-size four-element vector operations and scalar arithmetic, eliminating dynamic memory allocation. Trigonometric functions employ lookup tables or polynomial approximations to reduce latency; random sampling for discrete Lévy and von Mises distributions uses lightweight generators with precomputed probability tables to minimize normalization overhead. These techniques preserve constant-time operation across all control cycles. The Bitcraze Crazyflie platform (STM32F405 ARM Cortex-M4, 168 MHz, 192 kB SRAM) exemplifies the target hardware; the minimal state requirements (only pose and heading estimates) are well within its memory constraints, supporting operation at the typical update rates (10--20 Hz) of nano-UAV flight control.

\section{Experimental Setup}

We evaluate the proposed Sensor-Driven Lévy-Walk (SDLW) controller in the open-source \textbf{IR-SIM} simulator, a lightweight Python framework for navigation, control, and multi-robot scenarios with configurable robots, sensors, and environments \cite{ir-sim}. Evaluation focuses on coverage efficiency and collision robustness across three representative arenas and five team sizes. We compare against a baseline (Uniform Heading Lévy Walk) under identical sensing, kinematics, and safety constraints to isolate the contribution of sensor-adaptive heading selection.

\subsection{Simulator and Configuration}

\textbf{Simulator.} IR-SIM v2.9.0 provides collision detection, configurable holonomic kinematics, 1D time-of-flight range sensing, and scenario specification via YAML. Scenarios execute with physics integration and rendering disabled for high-throughput, repeatably randomized trials \cite{ir-sim}.

\begin{figure}[tbp]
\centering
\begin{tabular}{@{}c@{\hspace{6mm}}c@{}}
\includegraphics[width=0.41\columnwidth]{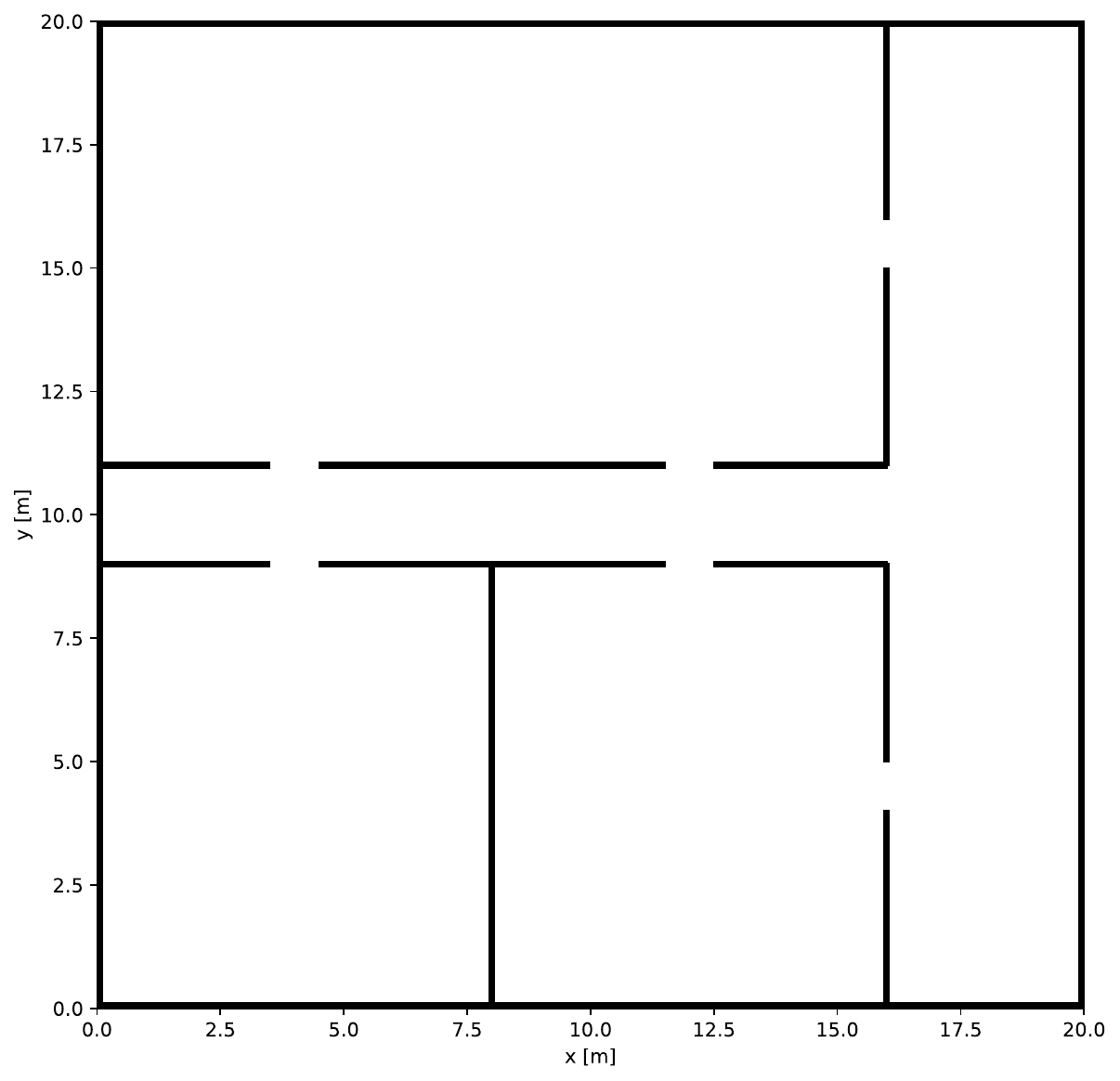} &
\includegraphics[width=0.41\columnwidth]{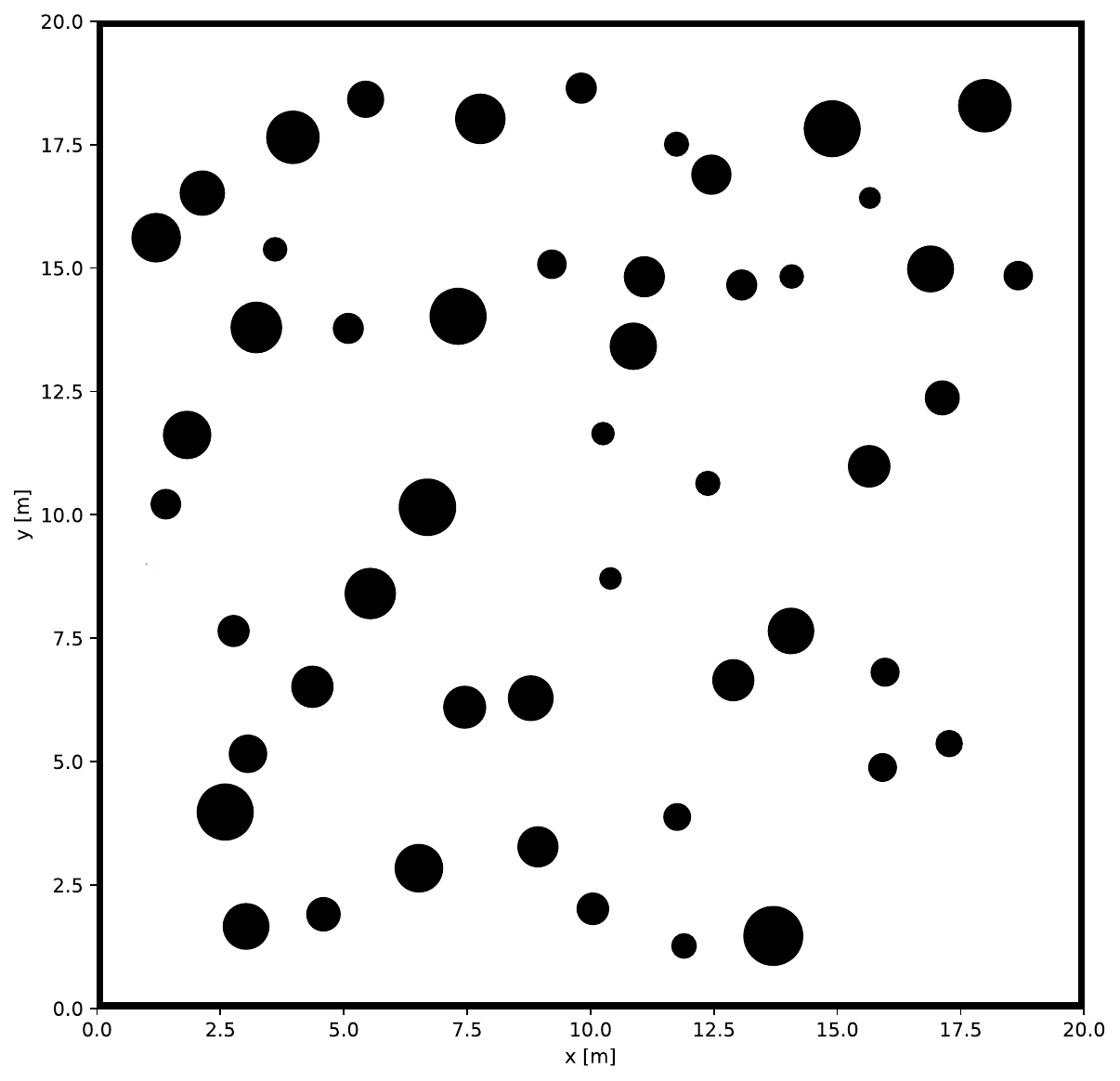} \\
(a) Rooms \& Corridors & (b) Cluttered (50)
\end{tabular}
\caption{Representative arena layouts: (a) three rooms connected by a 2~m-wide horizontal corridor and a 4~m-wide vertical corridor with 1 m door openings; (b) 50 randomly placed obstacles (seed 42 example).}
\label{fig:arenas}
\end{figure}

\textbf{Arenas (all 20 m $\times$ 20 m):} We evaluate on three environments representative of indoor nano-UAV operations: (1) \textit{Open}---rectangular boundary, no internal obstacles; (2) \textit{Rooms-and-corridors}---three rectangular rooms connected by a 2 m-wide horizontal corridor and a 4 m-wide vertical corridor with 1 m door openings; (3) \textit{Cluttered-50}---50 randomly placed non-overlapping circular obstacles with radii in [0.2, 0.6] m. These span from geometrically coherent (open/structured) to geometrically ambiguous (random), testing heading adaptation across structure diversity. Representative layouts for the structured and cluttered arenas are shown in Fig.~\ref{fig:arenas}.

\textbf{Robot model.} Holonomic kinematics with fixed forward speed \SI{0.5}{\meter\per\second}, saturated yaw rate $|\dot{\psi}|_{\max} = \SI{0.5}{\radian\per\second}$ (\SI{28.6}{\degree\per\second}), and collision radius \SI{0.10}{\meter}. These parameters reflect typical nano-UAV capabilities, with conservative speed and turn limits to allow safe operation in tight spaces and at low update rates.

\textbf{Sensing.} Four orthogonal 1D distance readings (front, left, right, back) in the agent's local frame, maximum range \SI{4}{\meter}, simulating the Bitcraze Crazyflie Multi-ranger deck.

\textbf{Parameters.} Step bounds $d_{\min}=\SI{0.5}{\meter}$, $d_{\max}=\SI{20}{\meter}$ span feasible segment lengths for sub-\SI{50}{\gram} platforms in \SI{20}{\meter}$\times$\SI{20}{\meter}{} arenas. Von Mises parameters $\kappa_{\min}=0.6$ and $\kappa_{\max}=10$ bound the concentration of the sampled headings. Backward penalty $\beta_B=0.3$, heading tolerance $\delta_\psi=\SI{5}{\degree}$, and obstacle threshold $r_{\mathrm{th}}=\SI{0.4}{\meter}$ use conservative safety margins suitable for 10--20~Hz onboard control.

\textbf{Baseline.} Uniform Heading Lévy Walk (UHLW): identical discrete Lévy steps and safety-handling as SDLW, but heading selected uniformly from $(-\pi,\pi)$ and the sensor-adaptive von~Mises branch is bypassed. This isolates the contribution of sensor-adaptive heading selection.

\textbf{Experimental protocol.} Team sizes: $k \in \{1,2,4,8,12\}$ agents. 50 independent random trials per condition, 15 minute horizon, grid-based coverage metric (0.25 m cells).

\section{Results}

We present results across three representative arenas and five team sizes. SDLW consistently outperforms UHLW, with gains correlated to environmental geometry coherence.

\subsection{Coverage Performance}

\begin{figure}[!t]
\centering
\textbf{Open arena}\\
\includegraphics[width=0.95\columnwidth]{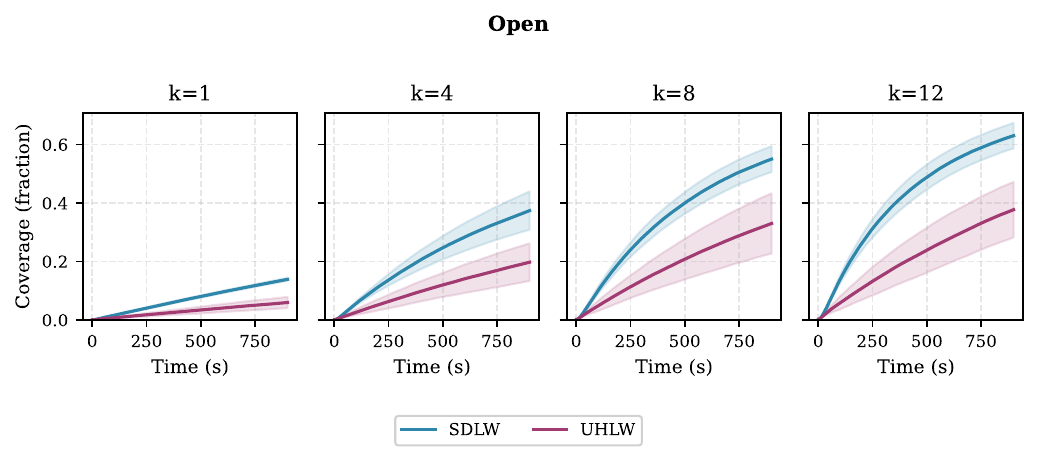}\\[4pt]
\textbf{Rooms \& Corridors}\\
\includegraphics[width=0.95\columnwidth]{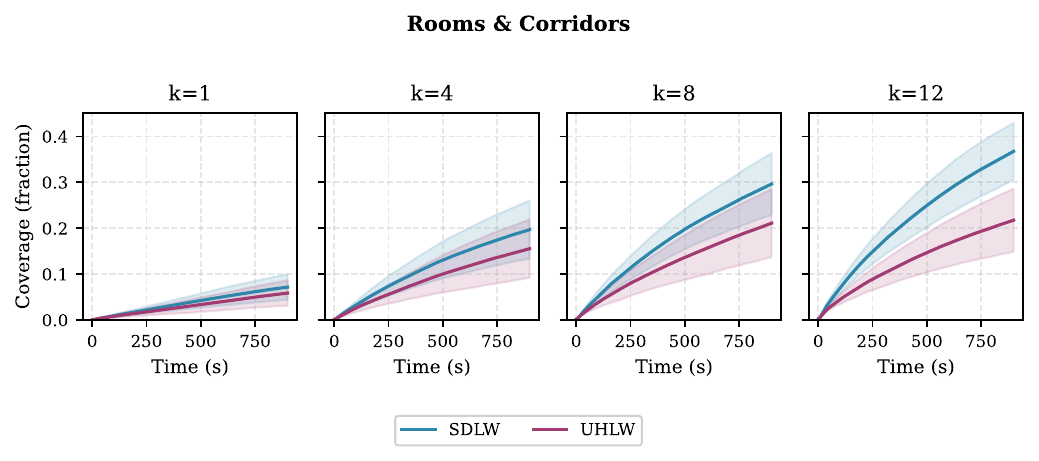}\\[4pt]
\textbf{Cluttered (50)}\\
\includegraphics[width=0.95\columnwidth]{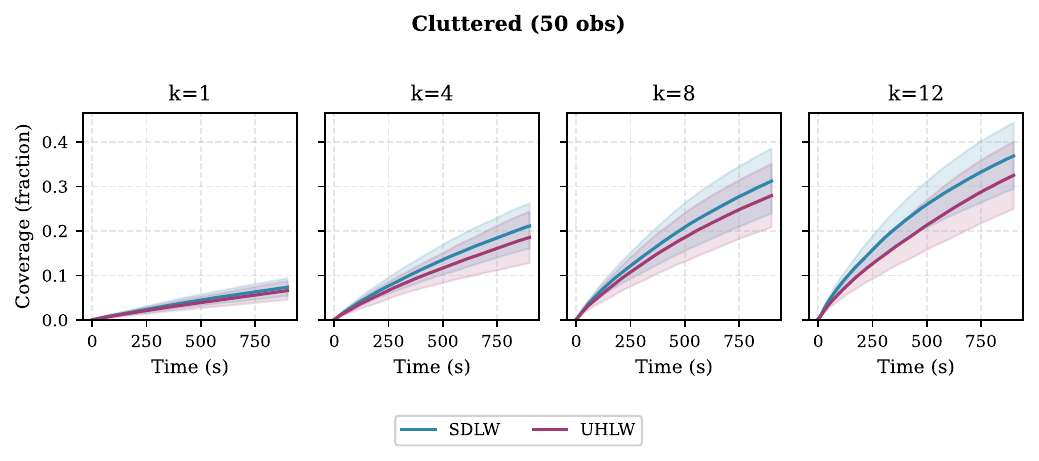}
\caption{Coverage over time for $k\in\{1,4,8,12\}$ agents (columns); results with $k=2$ omitted for brevity. Coverage is defined as the fraction of free-space grid cells (\SI{0.25}{\meter} resolution) visited by any team member. Mean coverage with standard-deviation shading over 50 trials. Blue: SDLW (sensor-adaptive); Purple: UHLW (uniform heading). SDLW substantially outperforms in structured environments, achieving 79.6\% higher coverage in Open and 43.1\% in Rooms \& Corridors, with 13.6\% advantage in cluttered layouts where sensor-adaptive heading benefits geometry-driven exploration.\\\\}
\label{fig:coverage-over-time}
\end{figure}

Fig.~\ref{fig:coverage-over-time} shows that SDLW substantially accelerates coverage accumulation relative to UHLW across all conditions, achieving 79.6\% higher mean coverage in Open, 43.1\% in Rooms and Corridors, and 13.6\% in Cluttered (50 obstacles) arenas consistently across all team sizes. Coverage gains are largest in structured environments and modest in random obstacles.

\subsection{Safety and Robustness}

\begin{figure}[!t]
\centering
\includegraphics[width=1.0\columnwidth]{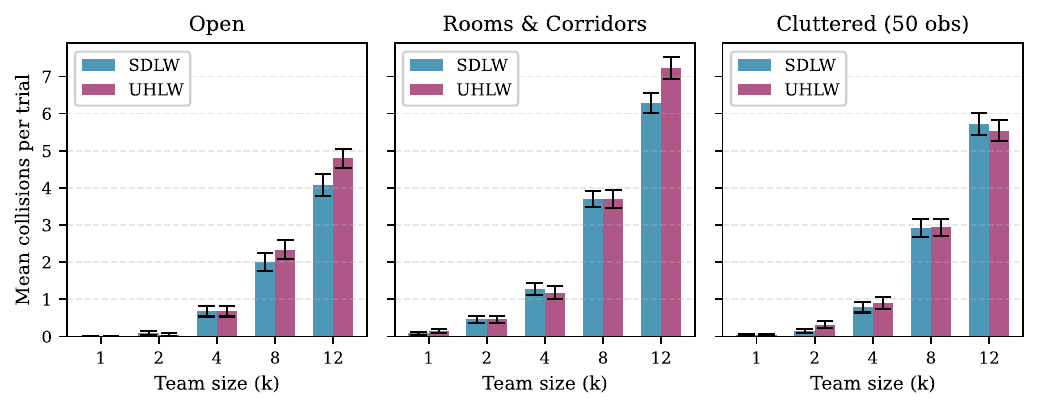}
\caption{Collision events per trial across arenas and team sizes. Mean collisions for SDLW (ours, blue) and Uniform Heading Lévy Walk (UHLW, purple), with error bars showing standard error. SDLW demonstrates consistent collision reduction across all environments: Open shows a particularly strong 13.0\% mean reduction, while Rooms \& Corridors achieves 7.1\% fewer collisions. Even in the cluttered arena, SDLW maintains a modest 1.4\% collision advantage. The sensor-adaptive strategy provides collision benefits across environments, with the magnitude varying based on geometric structure coherence.}
\label{fig:collisions}
\vspace{0.4\baselineskip}
\end{figure}

Fig.~\ref{fig:collisions} summarizes collision robustness, showing consistent SDLW advantages across environments. In the Open arena, SDLW incurs 13.0\% fewer mean collisions than UHLW across all team sizes, with the benefit particularly pronounced at larger $k$: at $k=12$, SDLW achieves 4.08 collisions compared to UHLW's 4.80. The Rooms and Corridors environment shows a moderate SDLW advantage, with 7.1\% lower mean collisions and consistent improvements across most team sizes tested.

In the Cluttered (50 obstacles) arena, SDLW maintains a modest collision advantage with 1.4\% fewer mean collisions than UHLW overall. At moderate team sizes ($k=8$), collision rates are comparable (2.92 vs. 2.94), while at $k=12$ collision counts show SDLW with 5.72 and UHLW with 5.54. Collision reduction benefits are substantial in open and structured spaces, and modest in random obstacle layouts. The current logging pipeline records aggregate per-robot collision flags rather than classifying contacts as obstacle or inter-agent events, so these counts should be interpreted as overall collision robustness rather than type-specific failure rates.

\subsection{Limitations}
Key limitations: (i)~simulation assumes perfect noiseless sensing; real sensors exhibit noise and measurement artifacts; (ii)~2-D planar model abstracts vertical dynamics and 3-D obstacles; (iii)~fixed parameters throughout execution; adaptive tuning could reduce parameter sensitivity by learning $\kappa$ values from brief exploration phases, at the cost of increased computational overhead; (iv)~hardware validation on actual nano-UAVs with real sensing and actuation latencies remains essential.

\section{Conclusion}

This work presents a Sensor-Driven Lévy-Walk (SDLW) controller for palm-sized nano-UAVs, combining discrete Lévy step lengths with sensor-adaptive von Mises heading selection. The controller operates at constant computational cost, requires only four directional distance readings, and enables decentralized exploration without inter-robot communication for exploration control or prior knowledge. Simulation results demonstrate substantial coverage gains (up to 79.6\%) and collision reduction (up to 13.0\%) from sensor-weighted directional bias, with benefits largest in structured environments where geometry provides strong directional cues. SDLW provides a practical, lightweight baseline for scalable multi-robot exploration suitable for resource-constrained teams. Future work includes online learning of concentration parameters $\kappa_{min}$, $\kappa_{max}$ and resultant-strength-to-concentration mappings, temporal smoothing of sensor histories to reduce noise sensitivity, consideration of ToF sensors with higher angular resolution, and hardware validation on physical nano-UAV platforms under real sensing and actuation constraints.

\bibliographystyle{IEEEtran}
\bibliography{references.bib}

\end{document}